\documentclass[10pt]{article}
\usepackage{fancyhdr}
\usepackage{extramarks}
\usepackage{amsmath}
\usepackage{amsthm}
\usepackage[utf8]{inputenc}   
\usepackage[T1]{fontenc}
\usepackage{newunicodechar}   
 \usepackage{dblfloatfix} 
\usepackage{float}

\usepackage{booktabs}
\usepackage{tabularx}

\renewcommand{\arraystretch}{1.2}  

\newunicodechar{α}{$\alpha$}
\newunicodechar{β}{$\beta$}
\newunicodechar{γ}{$\gamma$}
\newunicodechar{κ}{$\kappa$}
\newunicodechar{δ}{$\delta$}

\newunicodechar{–}{--}       
\newunicodechar{—}{---}      
\newunicodechar{’}{'}        
\newunicodechar{±}{$\pm$}

\newunicodechar{­}{}
\usepackage{amsfonts}
\usepackage{siunitx}
\usepackage{tikz}
\usepackage[plain]{algorithm}
\usepackage{algpseudocode}
\usepackage{multirow}
\usepackage{booktabs}
\usepackage{graphicx}
\usepackage{subfigure}
\usepackage[margin=1in]{geometry}
\usepackage{booktabs}
\usepackage{makecell}
\usepackage{array} 
\usepackage[colorlinks,linkcolor=black,anchorcolor=black,citecolor=black,urlcolor=blue]{hyperref}
\usepackage{hyphenat}
\usepackage{amsmath,bm}
\usepackage{booktabs}
\usepackage{mathtools}
\usepackage{amssymb}
\usepackage{tikz-cd}
\usepackage{caption}
\usepackage{capt-of}
\usepackage{mciteplus}
\usepackage{cite}
\usepackage{mathrsfs}
\usepackage[title,titletoc,toc]{appendix}
\usepackage{xr}
\usepackage{parskip}
\usepackage{soul}
\usepackage{textcomp}
\usepackage[colaction]{multicol}
\usepackage[switch]{lineno}
\usepackage{lipsum}
\usepackage{etoolbox}
\usepackage{longtable}
\usepackage{array}
\usepackage{tablefootnote}
\usepackage{ragged2e}
\usepackage{soul}
\newcolumntype{C}[1]{>{\centering\arraybackslash}p{#1}}
\usetikzlibrary{automata,positioning}
\usepackage{float}

\usetikzlibrary{automata,positioning}

\newtheorem{remark}{Remark}[section]

\usepackage{amsthm}  %
\theoremstyle{definition}
\newtheorem{definition}{Definition}[section]

\usepackage{xr}
\begin{document}
	\title{Multi-dimensional Persistent Sheaf Laplacians for Image Analysis}
    
	\author{Xiang Xiang Wang$^{1}$,  and Guo-Wei Wei$^{1,2,3}$\footnote{
			Corresponding author.		Email: weig@msu.edu} \\
		\\
		$^1$ Department of Mathematics, \\
		Michigan State University, East Lansing, MI 48824, USA.\\
        $^2$ Department of Biochemistry and Molecular Biology,\\
		Michigan State University, East Lansing, MI 48824, USA.  \\
		$^3$ Department of Electrical and Computer Engineering,\\
		Michigan State University, East Lansing, MI 48824, USA. \\
        \\
	}
	\date{\today} 
	
	\maketitle
	
\begin{abstract}
We propose a multi-dimensional persistent sheaf Laplacian (MPSL) framework on simplicial complexes for image analysis.  The proposed method is motivated by the strong sensitivity of commonly used dimensionality reduction techniques, such as principal component analysis (PCA), to the choice of reduced dimension. Rather than selecting a single reduced dimension or averaging results across dimensions, we exploit complementary advantages of multiple reduced dimensions. At a given dimension, image samples are regarded as simplicial complexes, and  persistent sheaf Laplacians are utilized to extract a multiscale localized topological spectral representation for individual image samples.  Statistical summaries of the resulting spectra are then aggregated across scales and dimensions  to form  multiscale multi-dimensional image representations.  We evaluate the proposed framework on the COIL20 and ETH80 image datasets using standard classification protocols. Experimental results show that the proposed method provides more stable performance across a wide range of reduced dimensions and achieves consistent improvements to PCA-based baselines in moderate dimensional regimes.
\end{abstract}


    
 \newpage
	 
\section{Introduction }
Dimensionality reduction is a fundamental problem in  data science, in which high-dimensional observations are mapped to lower-dimensional representations for tasks such as classification, clustering, and visualization \cite{fodor2002survey,bishop2006pattern}. In image analysis and image processing, effective feature extraction plays a particularly important role, as most machine learning and pattern recognition methods operate on feature representations rather than raw image data \cite{gonzalez2009digital}. To this end, a wide range of dimensionality reduction techniques have been developed and widely adopted, including classical linear methods such as principal component analysis (PCA)\cite{bro2014principal}, factorization-based approaches such as nonnegative matrix factorization (NMF) \cite{lee1999learning}, and nonlinear manifold learning methods such as UMAP \cite{mcinnes2018umap} and many related techniques. Among these methods, PCA remains a benchmark for other approaches\cite{zhou2022pca}. Despite their empirical success, these methods are known to be sensitive to the choice of reduced dimension. In practice, the choice of different dimensions can lead to substantially different representations and varying downstream performance. In practice, the optimal selection of dimensionality is often problem-dependent and   empirical \cite{burges2010dimension,hozumi2022ccp}. As a result, practitioners often rely on trial and error or empirical choices, raising concerns regarding the stability, robustness, and reproducibility of feature extraction pipelines. 

From the perspective of image feature representation, sensitivity to dimensional choice presents a fundamental challenge. Image data often exhibit complex structures across multiple scales \cite{burt1987laplacian}, and representations obtained at a single fixed dimension may not capture complementary information present at other resolutions \cite{maaten2008visualizing}. Selecting a single dimension risks discarding informative structural patterns, while averaging representations across dimensions may obscure scale-specific characteristics \cite{mallat2012group}. Consequently, developing feature extraction methods that are stable with respect to different dimension selections and capable of integrating information across multiple dimensional scales is an important and challenging problem in image analysis.

Topological data analysis (TDA) \cite{edelsbrunner2010computational,zomorodian2004computing} provides a principled framework for capturing intrinsic structural information in complex data. Paired with topological deep learning \cite{cang2017topologynet,papamarkou2024position},  TDA has found successful applications across a wide range of scientific domains, including molecular and materials modeling, biological data analysis, and dynamical systems, topological descriptors have been shown to capture structural and relational patterns in complex systems \cite{wee2025review}.

Persistent homology \cite{edelsbrunner2010computational,zomorodian2004computing} has been one of the most commonly used tools over years. It tracks the appearance and disappearance of topological binvariants across scales by constructing filtration on data representations, and provides multiscale multi-dimensional summaries that are stable under small perturbations. Persistent homology has been applied in a variety of settings. Persistent homology has been extensively studied and applied to image data, where it characterizes topological features across scales through filtration defined on image representations \cite{garin2019topological}. Nonetheless, persistent homology has many limitations, including the inability of capturing non-topological shape evolution over scales. Persistent spectral theory was introduced in terms of persistent combinatorial Laplacians (also know as persistent Laplacians) to deal with this limitation \cite{wang2020persistent}.  The kernel of persistent combinatorial Laplacians, i.e., their harmonic spectrum, gives rise to the same topological invariants as those given by persistent homology, while non-harmonic spectrum delivers additionally information about non-topological geometric changes over the filtration. Computational algorithms \cite{wang2021hermes, memoli2022persistent,jones2025petls} and stability analysis \cite{liu2024algebraic} have been developed for persistent Laplacians. A comprehensive test on more thirty datasets indicates  persistent Laplacians outperforms  persistent homology  \cite{qiu2023persistent}. More recently, some researchers found that persistent Laplacian is particularly suited to image analysis \cite{davies2023persistent}. Earlier persistent Laplacians were defined on simplicial complexes. Definitions on various other topological spaces have been developed, including persistent path Laplacians\cite{wang2023persistent}, persistent sheaf Laplacians \cite{wei2025persistent}, etc.
This topological spectral approach extends the scope and capability of TDA as discussed in recent survey \cite{wei2025persistent} and review \cite{su2025topological}. 

With an exception of manifold topological deep learning via de Rham-Hodge theory \cite{liu2025manifold}, most existing TDA-based approaches for image analysis rely on cubical complexes constructed directly from image grids, where each image is treated independently and topological or spectral features are extracted from pixel-level structures. While cubical constructions are convenient for grid-based data, they impose fixed local connectivity and limit flexibility in representing relationships beyond individual images, particularly higher-order interactions among images within a dataset.

 Simplicial complexes provide a more flexible alternative for modeling network relationships, as they naturally encode higher-order interactions among multiple entities. In TDA, simplicial constructions are widely used for point cloud and network data, but their use in image analysis has remained relatively limited. Existing image-based TDA methods rarely construct simplicial complexes at the dataset level, where image samples are treated as vertices and simplices are formed based on similarity or neighborhood relations among image samples. As a result, structural information arising from interactions among similar images is typically not explicitly captured in cubical-based image representations.

However, simplicial complexes on the dataset level alone do not work for many network analysis tasks, as one also needs localization. Sheaf-theoretic extensions of graph and simplicial Laplacians have been actively developed to model individual data entries in a dataset and their consistency across different regions \cite{curry2014sheaves,hansen2019toward}. Persistent sheaf Laplacians further extend these ideas by incorporating multiscale filtrations, enabling spectral information to be analyzed across scales \cite{wei2025persistent}. While sheaf Laplacians and their persistent variants have been explored in several application domains \cite{hayes2025persistent}, their potential for image analysis has not yet been systematically investigated. 

 From the perspective of dimensionality reduction and feature representation, image data often exhibit patterns at multiple resolutions. Representations obtained at a single fixed reduced dimension may fail to capture complementary information present at other dimensional scales, while selecting a single dimension in advance is typically problem-dependent and empirically determined \cite{li2020efficient,hozumi2022ccp}. This motivates treating different reduced dimensions not merely as hyperparameters to be tuned, but as complementary scales that capture distinct structural patterns. Integrating information across multiple dimensional resolutions therefore offers a natural way to improve stability and robustness in feature extraction. 

 A key distinction between the proposed framework and existing TDA-based image analysis methods lies in the level at which topological structures are constructed. While cubical complexes are typically built on pixel grids and treat each image independently, our approach constructs simplicial complexes at the dataset level, where images are viewed as nodes in a network and higher-order interactions are modeled through simplices.

Motivated by these observations, this paper investigates the use of simplicial complexes and persistent sheaf Laplacians for image feature extraction within a multi-dimensional persistent sheaf Laplacian (MPSL) framework. Image samples in a dataset are analyzed at multiple dimensions. At a given dimension, image samples are regarded as a point cloud.  A family of dataset-level simplicial complexes is created via filtration for each image sample based on its neighborhood relations among images. Sheaf Laplacians defined on these complexes enable local structural information to be coupled across images in the dataset through restriction maps. As such, persistent sheaf Laplacians provide access to spectral information. By further integrating spectral representations across multiple reduced dimensions, the proposed framework aims to extract features that are less sensitive to the choice of dimensionality  and more stable across scales.

We evaluate the proposed approach on standard image datasets, including COIL20 \cite{nene1996columbia} and ETH80 \cite{leibe2003analyzing} benchmark. Experimental results demonstrate that the multi-dimensional persistent sheaf Laplacian framework achieves stable performance across a wide range of reduced dimensions and, by integrating information across dimensions, consistently attains substantially higher classification accuracy than PCA-based baselines, including dimension-averaged PCA.

The remainder of this paper is organized as follows. Section~2 reviews the necessary background on simplicial complexes, sheaf Laplacians, and related spectral constructions.
Section~3 introduces the proposed multi-dimensional persistent sheaf Laplacian framework and describes how information is integrated across reduced dimensions and neighborhood scales.
Section~4 presents the experimental setup and reports the results of image classification experiments, together with a detailed discussion of the findings.
Section~5 provides further discussion and analysis of the proposed method.
Finally, Section~6 concludes the paper and outlines possible directions for future work.

\section{Mathematical Foundations}

This section introduces the mathematical concepts and notation
used throughout the paper.
We briefly review the theory of cellular sheaves on simplicial complexes,
the associated cochain complexes and sheaf Laplacians,
as well as filtrations and spectral operators.
These tools provide the formal foundation for the persistent sheaf Laplacian framework
and for the subspace-based representation used for comparison in later sections.
For convenience, Table~\ref{tab:notation} summarizes the main notation
used throughout the remainder of the paper.

\begin{table}[t]
\centering
\small
\setlength{\tabcolsep}{6pt}
\renewcommand{\arraystretch}{1.15}
\begin{tabular}{c|l}
\hline
\textbf{Symbol} & \textbf{Description} \\
\hline
$\mathcal{I}=\{I_i\}_{i=1}^m$ & Image dataset consisting of $m$ images \\

$I_i$ & The $i$-th image in the dataset \\

$p \times q$ & Spatial resolution of an image \\

$n=pq$ & Dimension of the vectorized image \\
$\mathbb{R}^n$ & Vector with dimension n\\
$\mathbb{R}^{m\times n}$ & Matrix with size $m\times n$\\
$x_i \in \mathbb{R}^{1\times n}$ & Vectorized representation of image $I_i$ \\

$X \in \mathbb{R}^{m\times n}$ & Data matrix formed by stacking all image vectors \\

$\mathcal{D}=\{d_1,\dots,d_r\}$ & Set of reduced dimensions \\

$d$ & A reduced feature dimension, $d\in\mathcal{D}$ \\

$X^{(d)} \in \mathbb{R}^{m\times d}$ & Feature matrix reduced to dimension $d$ \\

$x_i^{(d)} \in \mathbb{R}^{1\times d}$ & Reduced representation of $I_i$ at dimension $d$ \\

$\mathcal{K}=\{k_1,\dots,k_\ell\}$ & Set of neighborhood sizes \\

$k$ & Neighborhood size parameter, $k\in\mathcal{K}$ \\

$\mathcal{N}^{(d)}_k(I_i)$ & Set of $k$ nearest neighbors of $I_i$ in $\mathbb{R}^d$ \\

$\Sigma_i^{(d,k)}=\big(V_i^{(d,k)},\mathcal{S}_i^{(d,k)}\big)$ & Local simplicial complex centered at $I_i$ for $(d,k)$ \\

$V_i^{(d,k)}$ & Vertex set of $\Sigma_i^{(d,k)}$ \\

$\mathcal{S}_i^{(d,k)}$ & Simplicial set of $\Sigma_i^{(d,k)}$ \\

$D_i^{(d,k)} \in \mathbb{R}^{|V_i^{(d,k)}|\times |V_i^{(d,k)}|}$ & Local distance matrix on $V_i^{(d,k)}$ \\

$\mathcal{F}_i^{(d,k)}$ & Cellular sheaf defined on $\Sigma_i^{(d,k)}$ \\

$\rho_{\tau \to \sigma}$ & Restriction map for a face inclusion $\tau \subseteq \sigma$ \\

$\kappa(\cdot)$ & Distance-based kernel used to define restriction maps \\

$\sigma$ & Kernel scale parameter in $\kappa(\cdot)$ \\

$L_{i,h}^{(d,k)}$ & (Persistent) sheaf Laplacian matrix on $\Sigma_i^{(d,k)}$ of order $h$ \\

$h \in \{0,1\}$ & Sheaf Laplacian order used in this work \\

$\Lambda_{i,h}^{(d,k)}$ & Multiset of eigenvalues of $L_{i,h}^{(d,k)}$ \\

$\mathbf{f}_{i,h}^{(d,k)} \in \mathbb{R}^{s}$ & Vector of $s$ statistics extracted from $\Lambda_{i,h}^{(d,k)}$ \\

$s$ & Number of statistical descriptors extracted from each spectrum \\

$\mathbf{z}_i \in \mathbb{R}^{2s|\mathcal{D}||\mathcal{K}|}$ & Concatenated multiscale feature vector for image $I_i$ \\
\hline
\end{tabular}
\caption{Summary of main notations used throughout the paper.}
\label{tab:notation}
\end{table}

\subsection{Simplices and Simplicial Complexes}

Simplices and simplicial complexes are fundamental combinatorial objects
in computational topology.
They provide a discrete representation of higher-order relations
among finite sets of elements and form a standard framework
for many constructions in topological data analysis
\cite{zomorodian2004computing}.

Let $u_0,u_1,\dots,u_k$ be points in $\mathbb{R}^d$.
A point $x=\sum_{i=0}^k \lambda_i u_i$ is an \emph{affine combination} of the $u_i$
if $\sum_{i=0}^k \lambda_i = 1$.
The points $u_0,\dots,u_k$ are \emph{affinely independent} if any two affine combinations
\[
x=\sum_{i=0}^k \lambda_i u_i,
\qquad
y=\sum_{i=0}^k \mu_i u_i,
\]
satisfy $x=y$ only when $\lambda_i=\mu_i$ for all $i$.
Equivalently, $u_0,\dots,u_k$ are affinely independent if and only if
the vectors $u_i-u_0$ for $1\le i\le k$ are linearly independent.

The convex hull of $u_0,\dots,u_k$,
\[
\sigma=\mathrm{conv}\{u_0,u_1,\dots,u_k\},
\]
is called a \emph{$k$-simplex}.
A \emph{face} of a simplex $\sigma$
is the convex hull of a non-empty subset of $\{u_0,\dots,u_k\}$.

Using these notions, a simplicial complex can be defined as follows.

\begin{definition}[Simplicial Complex]
A \emph{simplicial complex} $K$ is a finite collection of simplices such that:
\begin{enumerate}
    \item if $\sigma \in K$ and $\tau \subseteq \sigma$, then $\tau \in K$;
    \item for any $\sigma, \tau \in K$, the intersection
    $\sigma \cap \tau$ is either empty or a common face of both.
\end{enumerate}
The set of all $k$-simplices in $K$ is denoted by $K^{(k)}$,
and the dimension of $K$ is defined as
\[
\dim K = \max_{\sigma\in K} \dim \sigma.
\]
\end{definition}

We write $\sigma \le \tau$ if $\sigma$ is a face of $\tau$.
A simplicial complex is a special case of a regular cell complex \cite{lundell2012topology,curry2014sheaves, hansen2019toward}.

\subsection{Sheaves on Simplicial Complexes and Sheaf Laplacians}

Cellular sheaves provide a flexible algebraic framework for encoding
local data and their consistency relations over a cell complex.
They were introduced and systematically developed in the context of
regular cell complexes, where they support a rich spectral theory
based on sheaf Laplacians
\cite{hansen2019toward}.

Simplicial complexes constitute a subclass of regular cell complexes,
and therefore naturally support the theory of cellular sheaves.
Beyond this structural compatibility, simplicial complexes arise
frequently in data-driven constructions, for instance from
pairwise distance matrices or neighborhood graphs.
This makes the simplicial setting particularly suitable for
computational and applied contexts. The formulation adopted here follows the simplicial framework
developed in \cite{wei2025persistent}, which extends sheaf Laplacian
theory to persistent settings via filtrations of simplicial complexes.

 \begin{definition}[Cellular Sheaf]
A \emph{cellular sheaf} $\mathcal{F}$ on a simplicial complex $X$ consists of the
following data:
\begin{enumerate}
    \item a simplicial complex $X$, where the face relation that $\sigma$ is a face
    of $\tau$ is denoted by $\sigma \le \tau$;
    \item an assignment of a finite-dimensional vector space $\mathcal{F}(\sigma)$
    to each simplex $\sigma$ of $X$, and a linear map
    \[
    \mathcal{F}_{\sigma\le\tau} : \mathcal{F}(\sigma) \to \mathcal{F}(\tau)
    \]
    to each face relation $\sigma \le \tau$,
    satisfying
    \[
    \rho \le \sigma \le \tau
    \quad\Longrightarrow\quad
    \mathcal{F}_{\rho\le\tau}
    =
    \mathcal{F}_{\sigma\le\tau} \circ \mathcal{F}_{\rho\le\sigma},
    \qquad
    \mathcal{F}_{\sigma\le\sigma} = \mathrm{id},
    \]
    where $\mathrm{id}$ denotes the identity map on $\mathcal{F}(\sigma)$.
\end{enumerate}
The vector space $\mathcal{F}(\sigma)$ is called the \emph{stalk} of $\mathcal{F}$
over $\sigma$, and the linear map $\mathcal{F}_{\sigma\le\tau}$ is called the
\emph{restriction map}.
\end{definition}


\subsection{Persistent Sheaf Laplacians}

Given a cellular sheaf on a simplicial complex, one can associate a cochain
complex whose algebraic structure is determined by the restriction maps of the
sheaf. This construction provides the foundation for defining sheaf Laplacians
and studying their spectral properties.

Let $\mathcal{F}$ be a cellular sheaf on a simplicial complex $X$.
For each integer $h \ge 0$, the space of \emph{$h$-cochains} with values in
$\mathcal{F}$ is defined as
\[
C^h(X;\mathcal{F})
=
\bigoplus_{\sigma \in X^{(h)}} \mathcal{F}(\sigma),
\]
that is, the direct sum of the stalks over all $h$-simplices of $X$.
An element of $C^h(X;\mathcal{F})$ assigns to each $h$-simplex $\sigma$
a vector in the stalk $\mathcal{F}(\sigma)$.

The \emph{coboundary operator}
\[
\delta_h : C^h(X;\mathcal{F}) \longrightarrow C^{h+1}(X;\mathcal{F})
\]
is defined using the restriction maps of the sheaf.
For each face relation $\sigma \le \tau$ with $\sigma \in X^{(h)}$
and $\tau \in X^{(h+1)}$, the corresponding block of $\delta_h$
is given by the restriction map
\[
\mathcal{F}_{\sigma \le \tau} : \mathcal{F}(\sigma) \to \mathcal{F}(\tau),
\]
up to a sign determined by a choice of orientation.
Collecting all such contributions yields a linear operator $\delta_h$
between the cochain spaces.

With this construction, the coboundary operators satisfy
\[
\delta_{h+1} \circ \delta_h = 0,
\]
and hence the sequence
\[
0 \longrightarrow C^0(X;\mathcal{F})
\stackrel{\delta_0}{\longrightarrow} C^1(X;\mathcal{F})
\stackrel{\delta_1}{\longrightarrow} C^2(X;\mathcal{F})
\longrightarrow \cdots
\]
forms a cochain complex, referred to as the \emph{sheaf cochain complex}.


When the simplicial complex is equipped with a filtration,
persistent variants of the sheaf Laplacian can be defined.
Let $\{X_t\}_{t \in \mathcal{T}}$ be a filtration of simplicial complexes,
that is, a nested family
\[
X_{t_1} \subseteq X_{t_2} \subseteq \cdots \subseteq X_{t_m},
\]
indexed by a finite ordered set $\mathcal{T}$.
Assume that a cellular sheaf $\mathcal{F}$ is defined on each $X_t$,
with compatible stalks and restriction maps across the inclusions.

For each scale parameter $t$ and each degree $h \ge 0$,
the sheaf cochain complex $(C^\bullet(X_t;\mathcal{F}), \delta_\bullet(t))$
induces a scale-dependent Laplacian operator defined by
\[
L_h(t)
=
\delta_h(t)^{\top}\delta_h(t)
+
\delta_{h-1}(t)\delta_{h-1}(t)^{\top},
\]
where $\delta_h(t)$ denotes the coboundary operator on $X_t$,
and the adjoint is taken with respect to the standard inner product
on the cochain spaces.
By convention, the second term is omitted when $h=0$.

For each fixed $t$, the operator $L_h(t)$ is symmetric and positive semidefinite,
and therefore admits a real, nonnegative spectrum.


To characterize features that persist across scales,
one may define operators associated with a filtration interval.
Let $a,b \in \mathcal{T}$ with $a \le b$.
The \emph{$h$-th $(a,b)$-persistent sheaf Laplacian}
is defined as an operator
\[
L^{a,b}_h : C^h(X_a;\mathcal{F}) \longrightarrow C^h(X_a;\mathcal{F}),
\]
given by
\[
L^{a,b}_h
=
\delta^{a,b}_h \, (\delta^{a,b}_h)^{\top}
+
\delta^{a}_{h-1} \, (\delta^{a}_{h-1})^{\top}.
\]

Here, $\delta^{a}_{h}$ denotes the coboundary operator on $X_a$,
and $\delta^{a,b}_h$ is the persistent coboundary operator induced by the
inclusion $X_a \hookrightarrow X_b$.
The adjoint is taken with respect to the standard inner product on the
cochain spaces.

The operator $L^{a,b}_h$ is symmetric and positive semidefinite.
Its spectrum encodes sheaf-consistent structures that remain compatible
with the inclusion $X_a \subseteq X_b$.
Different choices of $(a,b)$ yield a family of interval-dependent
persistent Laplacian operators.

\begin{remark}
Throughout the remainder of the paper, a fixed filtration interval $(a,b)$
is assumed. For notational simplicity, the superscript $(a,b)$
in the persistent sheaf Laplacian is suppressed.
\end{remark}

\subsection{Dimensionality Reduction and Multi-dimensional  Representations}

Dimensionality reduction is a fundamental component of many image analysis and machine learning pipelines, where high-dimensional observations are mapped to lower-dimensional representations for tasks such as classification, clustering, and visualization. Formally, given data points $x_i \in \mathbb{R}^D$, dimensionality reduction methods aim to construct a mapping
\[
\Phi_d : \mathbb{R}^D \rightarrow \mathbb{R}^d, \qquad d \ll D,
\]
where the reduced dimension $d$ controls the level of structural detail preserved in the representation.

Despite extensive study, there is generally no principled criterion to determine an optimal reduced dimension $d$ \emph{a priori}. In practice, $d$ is often selected empirically or fixed to a small set of representative values. Importantly, representations $\Phi_d(x)$ obtained at different values of $d$ are not simply nested or scaled versions of one another, but may encode qualitatively different geometric and statistical properties of the data.

Lower-dimensional embeddings tend to emphasize coarse global structure, while higher-dimensional embeddings preserve increasingly fine-scale variations at the cost of higher sensitivity to noise and redundancy. As a result, downstream performance, such as classification accuracy, can vary substantially with respect to the choice of $d$.

From this perspective, the reduced dimension $d$ can be interpreted as a scale parameter rather than a hyperparameter to be tuned or averaged over. Let $\mathcal{D} = \{ d_1, d_2, \ldots, d_m \}$ denote a set of reduced dimensions. Each mapping $\Phi_{d_k}$ provides a representation of the data at a distinct resolution, capturing complementary structural information. This observation motivates multiscale representations that integrate information across multiple reduced dimensions, with the goal of achieving stable and robust performance without committing to a single dimensional choice.



\section{A Multi-dimensional Persistent Sheaf Laplacian Framework for Image Analysis}
\label{sec:framework}

\begin{figure}[hbpt!]
\centering
\includegraphics[width=\linewidth]{ 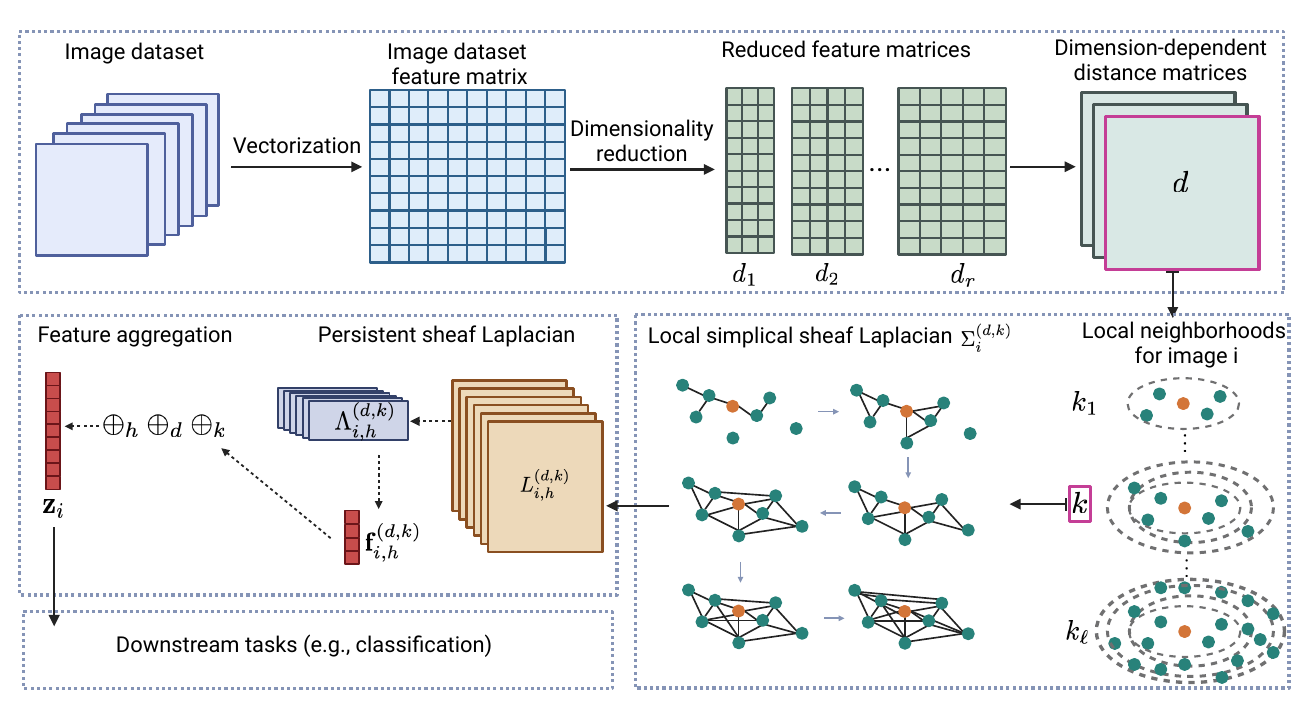}
\caption{
Overview of the proposed multi-dimensional persistent sheaf Laplacian framework.
Given an image dataset, each image is first vectorized
to form a feature matrix. Dimensionality reduction is then performed to obtain
multiple reduced feature matrices corresponding to different dimensions
$d \in \mathcal{D}$. For each reduced dimension, a dimension-dependent
distance matrix is computed.
For each image   and each pair $(d,k)$ with neighborhood size
$k \in \mathcal{K}$, a local simplicial complex
$\Sigma_i^{(d,k)}$ is constructed from the $k$-nearest neighborhood.
A sheaf structure is defined on each local complex, leading to
dimension- and scale-dependent sheaf Laplacian matrices
$L_{i,h}^{(d,k)}$ for $h \in \{0,1\}$.
Their spectra $\Lambda_{i,h}^{(d,k)}$ are tracked across filtration
scales to obtain persistent spectral information.
Statistical descriptors extracted from the eigenvalue sets are denoted
by $\mathbf{f}_{i,h}^{(d,k)}$ and aggregated across all dimensions and scales
to form the unified feature vector $\mathbf{z}_i$.
The resulting representation is used for downstream tasks,
such as classification.
}
\label{fig:framework_overview}
\end{figure}

In this section, we introduce a multi-dimensional image feature extraction framework based on persistent sheaf Laplacians defined on dataset-level simplicial complexes.
The proposed framework incorporates a multi-dimensional and multiscale structure.
The first level arises from varying the reduced dimension used to construct image representations, while the second level is induced by varying the neighborhood size used to define local simplicial structures.
Rather than selecting a single reduced dimension or a single neighborhood parameter, we treat different parameter settings as complementary scales and integrate the resulting spectral features within a unified representation.
The overall pipeline of the proposed framework is illustrated in Figure~\ref{fig:framework_overview}.

\subsection{Overview of the Multi-dimensional persistent sheaf Laplacian   framework}
\label{subsec:framework_overview}

Let $\mathcal{I} = \{ I_i \}_{i=1}^m$ denote an image dataset consisting of $m$ images.
Each image $I_i$ with size $p \times q$ is represented as a matrix in $\mathbb{R}^{p \times q}$.
By vectorizing each image in a row-wise manner, we obtain a feature vector
\[
x_i \in \mathbb{R}^{1 \times n}, \quad n = pq.
\]
Stacking all image vectors yields the feature matrix
\[
X =
\begin{bmatrix}
x_1 \\
x_2 \\
\vdots \\
x_m
\end{bmatrix}
\in \mathbb{R}^{m \times n},
\]
where each row corresponds to one image in the dataset.

\medskip
\noindent\textbf{Step 1 (Dimensionality reduction).}
To mitigate the high dimensionality of the original feature space, we apply a dimensionality reduction method to $X$.
Rather than fixing a single reduced dimension, we consider a set of reduced dimensions
\[
\mathcal{D} = \{ d_1, d_2, \dots, d_r \}, \quad d_j \ll n.
\]
For each $d \in \mathcal{D}$, dimensionality reduction produces a reduced feature matrix
\[
X^{(d)} =
\begin{bmatrix}
x_1^{(d)} \\
x_2^{(d)} \\
\vdots \\
x_m^{(d)}
\end{bmatrix}
\in \mathbb{R}^{m \times d},
\]
where $x_i^{(d)} \in \mathbb{R}^{1 \times d}$ denotes the reduced representation of image $I_i$.
Each reduced dimension is regarded as providing a complementary representation of the dataset.

\medskip
\noindent\textbf{Step 2 (Dataset-level simplicial complex construction).}
For a fixed reduced dimension $d \in \mathcal{D}$, the vectors $\{x_i^{(d)}\}_{i=1}^m$ are treated as vertices in a dataset-level simplicial representation.
To capture local geometric relationships, we consider a set of neighborhood sizes \cite{lesnick2015interactive}
\[
\mathcal{K} = \{ k_1, k_2, \dots, k_\ell \}.
\]
For each $k \in \mathcal{K}$, a $k$-nearest neighbor graph is constructed based on the vectors $\{x_i^{(d)}\}_{i=1}^m$.
Local simplicial complexes are then formed around each image by including simplices induced by its $k$-nearest neighbors.

\noindent\textbf{Step 3 (Persistent sheaf Laplacian construction).}
For each image $I_i$, each reduced dimension $d \in \mathcal{D}$, and each neighborhood size $k \in \mathcal{K}$,
we construct a local simplicial complex
\[
\Sigma_i^{(d,k)} = (V_i^{(d,k)}, \mathcal{S}_i^{(d,k)}),
\]
where the vertex set $V_i^{(d,k)}$ consists of the image $I_i$ together with its $k$ nearest neighbors
in the reduced feature space $\mathbb{R}^d$.
The simplicial set $\mathcal{S}_i^{(d,k)}$ encodes higher-order relations induced by these neighborhood
connections, for instance via clique expansion of the corresponding $k$-nearest neighbor graph.

On each local simplicial complex $\Sigma_i^{(d,k)}$, we define a sheaf structure by assigning local data
to simplices and specifying restriction maps that encode consistency relations across adjacent simplices.
For each Laplacian order $h \in \{0,1\}$, we construct the corresponding sheaf Laplacian matrix
\[
L_{i,h}^{(d,k)}
\]
This matrix encodes the sheaf-consistent interactions between $h$-simplices
within the local complex $\Sigma_i^{(d,k)}$.

To incorporate persistence, the construction is performed across a filtration
parameter, and the spectral information of the sheaf Laplacians
$L_{i,h}^{(d,k)}$ is tracked across filtration scales,
yielding persistent spectral representations.

\medskip
\noindent\textbf{Step 4 (Statistical feature extraction).}
For each image $I_i$, reduced dimension $d \in \mathcal{D}$, neighborhood size $k \in \mathcal{K}$,
and sheaf Laplacian order $h \in \{0,1\}$,
we compute the eigenvalues of the corresponding sheaf Laplacian matrix
$L_{i,h}^{(d,k)}$.
Let
\[
\Lambda_{i,h}^{(d,k)}
=
\{ \lambda_{i,h,1}^{(d,k)}, \lambda_{i,h,2}^{(d,k)}, \dots,
   \lambda_{i,h,m_{i,h}^{(d,k)}}^{(d,k)} \}
\]
denote the multiset of eigenvalues of $L_{i,h}^{(d,k)}$,
where $m_{i,h}^{(d,k)}$ is the dimension of the matrix.

From each eigenvalue set $\Lambda_{i,h}^{(d,k)}$,
we extract a fixed collection of statistical descriptors,
including quantities such as the maximum eigenvalue,
the sum of eigenvalues,
the number of zero eigenvalues,
and the smallest nonzero eigenvalue.
Let $s$ denote the number of statistics extracted from each eigenvalue set.

\medskip
\noindent\textbf{Step 5 (Multi-dimensional feature integration and downstream tasks).}
For each image $I_i$, reduced dimension $d \in \mathcal{D}$, neighborhood size $k \in \mathcal{K}$,
and sheaf Laplacian order $h \in \{0,1\}$, let
\[
\mathbf{f}_{i,h}^{(d,k)} \in \mathbb{R}^{s}
\]
denote the vector of $s$ statistical descriptors extracted from the eigenvalue multiset
$\Lambda_{i,h}^{(d,k)}$.

We aggregate these descriptors across all $(d,k)$ and both Laplacian orders by concatenation:
\[
\mathbf{z}_i
=
\oplus_{h\in\{0,1\}}
\oplus_{d\in\mathcal{D}}
\oplus_{k\in\mathcal{K}}
\mathbf{f}_{i,h}^{(d,k)}
\;\in\;
\mathbb{R}^{2s|\mathcal{D}||\mathcal{K}|},
\]
where $\oplus$ denotes vector concatenation.

The resulting feature vector $\mathbf{z}_i$ is then used for downstream tasks such as classification,
for example via a $k$-nearest neighbor classifier under a fixed evaluation protocol.

These multi-dimensional representations are subsequently used as input to downstream learning tasks,
such as image classification.
By integrating information across both reduced dimensions and neighborhood scales,
the proposed framework avoids reliance on a single parameter choice and yields image representations
that are stable across a wide range of scales.

 \subsection{Dataset-level Simplicial Complex Construction}
\label{subsec:simplicial_construction}

We construct simplicial complexes on an image network defined at the dataset level,
where each vertex corresponds to an image and edges encode similarity relations
in a reduced feature space.
Unlike pixel-based constructions, this representation treats images as nodes of a shared network,
allowing interactions among images to be explicitly modeled.

Recall that for each reduced dimension $d \in \mathcal{D}$,
the image dataset is represented by a reduced feature matrix
$X^{(d)} \in \mathbb{R}^{m \times d}$,
whose $i$-th row $x_i^{(d)}$ corresponds to image $I_i$.
Based on these reduced representations, we construct simplicial complexes
that capture neighborhood relationships among images at the dataset level.

\medskip
\noindent\textbf{Distance and neighborhood graph.}
For a fixed reduced dimension $d$, we define a pairwise distance between images
in the reduced feature space, for example using the Euclidean distance
\[
\mathrm{dist}^{(d)}(i,j) = \| x_i^{(d)} - x_j^{(d)} \|_2 .
\]
These distances induce an image network through a $k$-nearest neighbor ($k$-NN) graph,
constructed for each neighborhood size $k \in \mathcal{K}$.
In this graph, each vertex represents an image, and an edge is placed between two vertices
if one image lies among the $k$ nearest neighbors of the other.
The resulting graph provides a scale-dependent description of local similarity structure
within the image dataset.

\medskip
\noindent\textbf{Local simplicial complexes.}
Rather than forming a single global simplicial complex for the entire dataset,
we focus on local simplicial structures centered at individual images.
For a fixed image $I_i$, reduced dimension $d \in \mathcal{D}$, and neighborhood size $k \in \mathcal{K}$,
we define a local vertex set
\[
V_i^{(d,k)} = \{ I_i \} \cup \mathcal{N}_k^{(d)}(I_i),
\]
where $\mathcal{N}_k^{(d)}(I_i)$ denotes the set of $k$ nearest neighbors of $I_i$
in the reduced feature space $\mathbb{R}^d$.

Using the induced subgraph of the $k$-NN graph on $V_i^{(d,k)}$,
we construct a local simplicial complex
\[
\Sigma_i^{(d,k)} = \big( V_i^{(d,k)}, \mathcal{S}_i^{(d,k)} \big),
\]
where the simplicial set $\mathcal{S}_i^{(d,k)}$ consists of all simplices
induced by neighborhood relations.
In practice, this can be achieved through clique expansion of the induced graph,
so that any set of mutually adjacent vertices forms a simplex.
Through this construction, higher-order simplices encode interactions among groups
of nearby images, rather than pairwise relations alone.

\begin{remark}
This dataset-level simplicial construction differs fundamentally from the cubical complexes
commonly used in image-based TDA, where simplices are defined on pixel grids
and each image is processed independently.
Here, images are treated as vertices of a shared image network,
and simplices reflect similarity relations among images in the dataset.
As a result, the local complexes $\Sigma_i^{(d,k)}$ adapt naturally to both
the reduced dimension $d$ and the neighborhood scale $k$,
providing a flexible geometric representation that supports multiscale multi-dimensional analysis.
\end{remark}

These local simplicial complexes form the geometric foundation for the
persistent sheaf Laplacian constructions introduced in the next subsection.

\subsection{Sheaf construction with distance-based restriction maps}
\label{subsec:psl_sheaf}

In this subsection, we describe how persistent sheaf Laplacians are defined
on the local simplicial complexes constructed in
Section~\ref{subsec:simplicial_construction},
and how spectral information is extracted for subsequent feature construction.
All indices are retained explicitly to reflect the multiscale multi-dimensional structure
induced by reduced dimensions and neighborhood sizes.

Fix an image $I_i$, a reduced dimension $d \in \mathcal{D}$, and a neighborhood
size $k \in \mathcal{K}$.
Let
\[
\Sigma_i^{(d,k)} = \big( V_i^{(d,k)}, \mathcal{S}_i^{(d,k)} \big)
\]
denote the local simplicial complex constructed in
Section~\ref{subsec:simplicial_construction}, and let
\[
D_i^{(d,k)} \in \mathbb{R}^{|V_i^{(d,k)}| \times |V_i^{(d,k)}|}
\]
be the corresponding local distance matrix, where
$D_i^{(d,k)}(u,v)$ denotes the distance between vertices $u$ and $v$
in the reduced feature space $\mathbb{R}^d$.

We equip $\Sigma_i^{(d,k)}$ with a cellular sheaf
$\mathcal{F}_i^{(d,k)}$.
In our construction, a one-dimensional real vector space is assigned
to every simplex:
\[
\mathcal{F}_i^{(d,k)}(\sigma) = \mathbb{R},
\quad \forall \, \sigma \in \mathcal{S}_i^{(d,k)}.
\]
As a consequence, each restriction map associated with a face inclusion
$\tau \subseteq \sigma$ is a linear map between one-dimensional vector spaces,
and therefore reduces to multiplication by a scalar.

The restriction maps are defined using local geometric information encoded
by the distance matrix $D_i^{(d,k)}$.
Let
\[
\kappa(t) = \exp\!\left( - \frac{t^2}{\sigma^2} \right)
\]
denote an exponential kernel with scale parameter $\sigma > 0$.
For a vertex-to-edge inclusion $\{u\} \subseteq \{u,v\}$, the restriction map
is defined as
\[
\rho_{\{u\} \to \{u,v\}}
=
\kappa\!\big( D_i^{(d,k)}(u,v) \big).
\]
For an edge-to-triangle inclusion $\{u,v\} \subseteq \{u,v,w\}$, the restriction
map is defined by averaging the contributions from the two incident vertices:
\[
\rho_{\{u,v\} \to \{u,v,w\}}
=
\frac{1}{2}
\Big(
\kappa\!\big( D_i^{(d,k)}(u,w) \big)
+
\kappa\!\big( D_i^{(d,k)}(v,w) \big)
\Big).
\]
All other restriction maps not involved in the construction are set to the
identity.

This choice yields a geometry-driven sheaf structure in which the coupling
between adjacent simplices is determined solely by distances in the reduced
feature space.
By using one-dimensional stalks and distance-based scalar restriction maps,
the resulting sheaf Laplacian captures local geometric consistency while
keeping the model simple and free of additional semantic or label-dependent
parameters.

\begin{remark}[Choice of the kernel scale parameter]
\label{rem:sigma_choice}
Unless otherwise stated, the kernel scale parameter $\sigma$ used in the
restriction maps is determined adaptively from local data.
Specifically, for each local simplicial complex $\Sigma_i^{(d,k)}$,
$\sigma$ is set to the median of all nonzero pairwise distances in the
corresponding local distance matrix $D_i^{(d,k)}$.
If no nonzero distances are present, $\sigma$ is set to $1$ by default.
This choice follows the implementation used in our experiments and avoids
introducing an additional hyperparameter, allowing the restriction maps to
adapt naturally to the local geometric scale.
\end{remark}

 \subsection{Statistical feature construction from persistent spectra}
\label{subsec:statistical_features}

For each image $I_i$, reduced dimension $d \in \mathcal{D}$, neighborhood size
$k \in \mathcal{K}$, and Laplacian order $h \in \{0,1\}$, the persistent sheaf
Laplacian construction yields a finite multiset of eigenvalues
\[
\Lambda_{i,h}^{(d,k)} =
\big\{ \lambda_{i,h,1}^{(d,k)}, \lambda_{i,h,2}^{(d,k)}, \dots \big\}.
\]
These eigenvalues encode spectral information associated with the local
simplicial structure and the sheaf-induced coupling at the scale $(d,k)$.

To obtain fixed-length feature representations suitable for downstream learning
tasks, each eigenvalue multiset $\Lambda_{i,h}^{(d,k)}$ is summarized using a
collection of real-valued statistical descriptors.
Each descriptor is defined as a function
\[
\phi : \mathcal{P}(\mathbb{R}_{\ge 0}) \rightarrow \mathbb{R},
\]
where $\mathcal{P}(\mathbb{R}_{\ge 0})$ denotes the collection of finite multisets
of nonnegative real numbers.

The statistics used in this work are listed in
Table~\ref{tab:psl_statistics}, together with their mathematical definitions and
brief interpretations.

\begin{table}[t]
\centering
\caption{Statistical descriptors computed from each persistent sheaf Laplacian
spectrum $\Lambda_{i,h}^{(d,k)}$.}
\label{tab:psl_statistics}
\begin{tabular}{lll}
\hline
Statistic & Definition & Interpretation \\ \hline
Zero eigenvalue count
& $\phi_{\mathrm{zero}}(\Lambda) = \#\{\lambda \in \Lambda : \lambda = 0\}$
& Kernel dimension \\

Smallest nonzero eigenvalue
& $\phi_{\min^+}(\Lambda) = \min\{\lambda \in \Lambda : \lambda > 0\}$
& Spectral gap \\

Maximum eigenvalue
& $\phi_{\max}(\Lambda) = \max \Lambda$
& Largest spectral value \\

Sum of eigenvalues
& $\phi_{\mathrm{sum}}(\Lambda) = \sum_{\lambda \in \Lambda} \lambda$
& Total spectral magnitude \\

Mean
& $\phi_{\mathrm{mean}}(\Lambda) = \frac{1}{|\Lambda|} \sum_{\lambda \in \Lambda} \lambda$
& Average value \\

Median
& $\phi_{\mathrm{med}}(\Lambda) = \mathrm{median}(\Lambda)$
& Central tendency \\

Standard deviation
& $\phi_{\mathrm{std}}(\Lambda) = \sqrt{\mathrm{Var}(\Lambda)}$
& Spectral dispersion \\

\hline
\end{tabular}
\end{table}

Among these descriptors, the number of zero eigenvalues
$\phi_{\mathrm{zero}}(\Lambda_{i,h}^{(d,k)})$ reflects the dimension of the kernel
of the corresponding sheaf Laplacian, while the smallest nonzero eigenvalue
$\phi_{\min^+}(\Lambda_{i,h}^{(d,k)})$ captures the spectral gap and is commonly
associated with the strength of coupling in Laplacian-based operators.
The remaining statistics summarize the overall scale and distribution of the
eigenvalues and provide complementary information about the spectral structure.

All statistics in Table~\ref{tab:psl_statistics} are computed independently for
each eigenvalue set $\Lambda_{i,h}^{(d,k)}$.
The resulting scalar values are concatenated across all reduced dimensions
$d \in \mathcal{D}$, neighborhood sizes $k \in \mathcal{K}$, and Laplacian orders
$h \in \{0,1\}$ to form the final multiscale multi-dimensional feature vector associated with image
$I_i$.

 \section{Experiments}
\label{sec:experiments}

In this section, we evaluate the proposed multi-dimensional persistent sheaf Laplacian (MPSL)
framework  on standard image classification benchmarks.
Our experiments focus on two widely used image datasets, COIL20 \cite{nene1996columbia} and ETH80 \cite{leibe2003analyzing},
which differ in visual complexity and intra-class variability.
The experimental design aims to assess both the stability of the proposed
features across different parameter settings and their effectiveness for
image classification, in comparison with PCA-based baselines.

\subsection{Datasets and experimental setup}
\label{subsec:datasets_setup}

This subsection describes the datasets, preprocessing steps, feature
representations, and experimental protocols used to evaluate the proposed
framework.
All methods are tested under identical input representations, parameter
ranges, and evaluation settings to ensure a fair and consistent comparison.

We conduct experiments on two standard image classification datasets,
COIL20  and ETH80, which are widely used in the image analysis literature and
exhibit different levels of visual complexity and intra-class variability.
The COIL20 dataset consists of grayscale images from $20$ object categories.
Each object is captured from $72$ uniformly sampled viewpoints by rotating it
on a turntable under controlled imaging conditions.
The background is uniform and illumination is stable, making COIL20 suitable
for evaluating representations under moderate viewpoint variation.
The ETH80 dataset contains images from $8$ object categories, with multiple
object instances per category, where each instance is imaged from several
viewpoints.
Compared to COIL20, ETH80 exhibits larger intra-class variability due to
differences among object instances and is therefore considered a more
challenging benchmark for image classification.

In our experiments, all color images in ETH80 are converted to grayscale prior
to feature extraction.
In addition, all images in both datasets are resized to a resolution of
$128 \times 128$ pixels.
These preprocessing steps ensure consistent input representations across
datasets and allow the subsequent feature extraction and simplicial
construction procedures to be applied uniformly.

\begin{table}[ht]
\centering
\caption{Summary of image datasets used in the experiments.}
\label{tab:datasets}
\begin{tabular}{lccccc}
\hline
Dataset & Classes & Images & Image type & Resolution & Characteristics \\ \hline
COIL20 \cite{nene1996columbia} & 20 & 1440 & Grayscale & $128 \times 128$ &
Single object per class \\
ETH80 \cite{leibe2003analyzing}  & 8  & 3280 & Grayscale (converted) & $128 \times 128$ &
Multiple object instances per class \\
\hline
\end{tabular}
\end{table}

Each image is vectorized into a high-dimensional feature vector before
dimensionality reduction is applied.
Rather than fixing a single reduced dimension, we consider a set of reduced
dimensions
\[
\mathcal{D} = \{200, 300, 400, 500,600,700,800,900, 1000\},
\]
and treat each choice as a distinct scale.
This design allows us to examine the sensitivity of classification performance
to the choice of reduced dimension and to construct multiscale multi-dimensional representations
by integrating information across different dimensional settings.

For each reduced dimension $d \in \mathcal{D}$, neighborhood relations among
images are defined using a $k$-nearest neighbor graph in the reduced feature
space.
We consider a range of neighborhood sizes
\[
\mathcal{K} = \{5,7, 10,12, 15,17, 20, 25, 30, 35, 40, 45, 50, 55, 60, 70, 80, 90, 100, 110\},
\]
which induces local simplicial complexes at different neighborhood scales.
Together, the reduced dimensions and neighborhood sizes define a two-level
multiscale structure that is used consistently throughout the experiments.

Persistent sheaf Laplacian spectra are computed on the local simplicial
complexes constructed for each pair $(d,k) \in \mathcal{D} \times \mathcal{K}$.
Statistical descriptors are then extracted from the spectra as described in
Section~\ref{sec:framework}, yielding fixed-length multiscale  multi-dimensional  feature vectors
for each image.

For classification, we employ a $k$-nearest neighbor classifier with
$k_{\mathrm{NN}} = 5$.
All methods are evaluated using $5$-fold cross-validation, with shuffling
enabled and a fixed random seed to ensure reproducibility.
The same classifier, parameter settings, and evaluation protocol are applied
to all methods, including PCA-based baselines and the proposed multi-dimensional
persistent sheaf Laplacian representations.
Classification performance is evaluated using accuracy (Acc), Macro Recall (MR),
and  Macro-F1 \cite{lewis1991evaluating, sokolova2009systematic, hand2001simple}.

\subsection{Results}
\label{subsec:results}

In this subsection, we report and analyze the classification performance
of MPSL framework on the COIL20 and ETH80 datasets.
All results are evaluated using Accuracy (Acc), Macro Recall (MR),
and  Macro F1, under the same experimental
protocol described in Section~\ref{subsec:datasets_setup}.
Unless otherwise stated, all comparisons are conducted using a
$k$-nearest neighbor classifier with $k_{\mathrm{NN}}=5$
and $5$-fold cross-validation.
\subsubsection{COIL20}
\label{subsubsec:coil20_results}

This experiment investigates the sensitivity of PCA-based representations to the choice of reduced dimension and evaluates whether the proposed MPSL representation can provide stable and robust performance in a high-dimensional regime.
We consider PCA dimensions ranging from 200 to 1000.
For each PCA dimension, PCA is used as a baseline representation.
For MPSL, spectral features are computed at the same fixed PCA dimension using multiple neighborhood sizes, and the resulting statistics are concatenated into a single feature vector for classification.
We refer to this setting as MPSL (single dimension $+$ multi-$k$).

Table~\ref{tab:coil20_pca_vs_psl_one} reports the classification results for PCA and MPSL (single dimension $+$ multi-$k$) across PCA dimensions 200 to 1000.
Performance is evaluated using Accuracy (Acc), Macro Recall (MR), and Macro-averaged F1 score (Macro F1).
For the PCA baseline, Acc and MR coincide in this evaluation protocol, and Macro F1 is additionally reported.

\begin{table*}[htbp!]
\centering
\small
\setlength{\tabcolsep}{5pt}
\renewcommand{\arraystretch}{1.15}
\caption{COIL20 results for PCA and MPSL (single dimension plus multi-$k$) over PCA dimensions 200--1000.}
\label{tab:coil20_pca_vs_psl_one}
\begin{tabular}{c|ccc|ccc}
\hline
PCA dim
& \multicolumn{3}{c|}{PCA}
& \multicolumn{3}{c}{MPSL } \\
\cline{2-7}
& Acc & MR & Macro F1
& Acc & MR & Macro F1 \\
\hline
200  & 0.7604 & 0.7604 & 0.7692 & 0.9049 & 0.9050 & 0.9039 \\
300  & 0.5632 & 0.5632 & 0.5956 & 0.9000 & 0.9003 & 0.8985 \\
400  & 0.4396 & 0.4396 & 0.4689 & 0.9146 & 0.9149 & 0.9133 \\
500  & 0.3854 & 0.3854 & 0.4019 & 0.9125 & 0.9130 & 0.9108 \\
600  & 0.3410 & 0.3410 & 0.3443 & 0.9118 & 0.9119 & 0.9098 \\
700  & 0.2688 & 0.2688 & 0.2547 & 0.9111 & 0.9112 & 0.9094 \\
800  & 0.2104 & 0.2104 & 0.1855 & 0.9118 & 0.9119 & 0.9098 \\
900  & 0.1785 & 0.1785 & 0.1570 & 0.9118 & 0.9120 & 0.9101 \\
1000 & 0.1604 & 0.1604 & 0.1368 & 0.9132 & 0.9133 & 0.9114 \\
\hline
Average & 0.3675 & 0.3675 & 0.3682 & 0.9091 & 0.9093 & 0.9074 \\
\hline
\end{tabular}
\end{table*}

As shown in Table~\ref{tab:coil20_pca_vs_psl_one}, the performance of the PCA baseline degrades substantially as the reduced dimension increases from 200 to 1000.
In contrast, MPSL (single dimension $+$ multi-$k$) achieves consistently high performance across all tested dimensions, with only minor variation.
On average, MPSL consistently outperforms PCA across all three evaluation metrics in this high-dimensional regime.

\begin{figure}[ht]
\centering
\includegraphics[width=\linewidth]{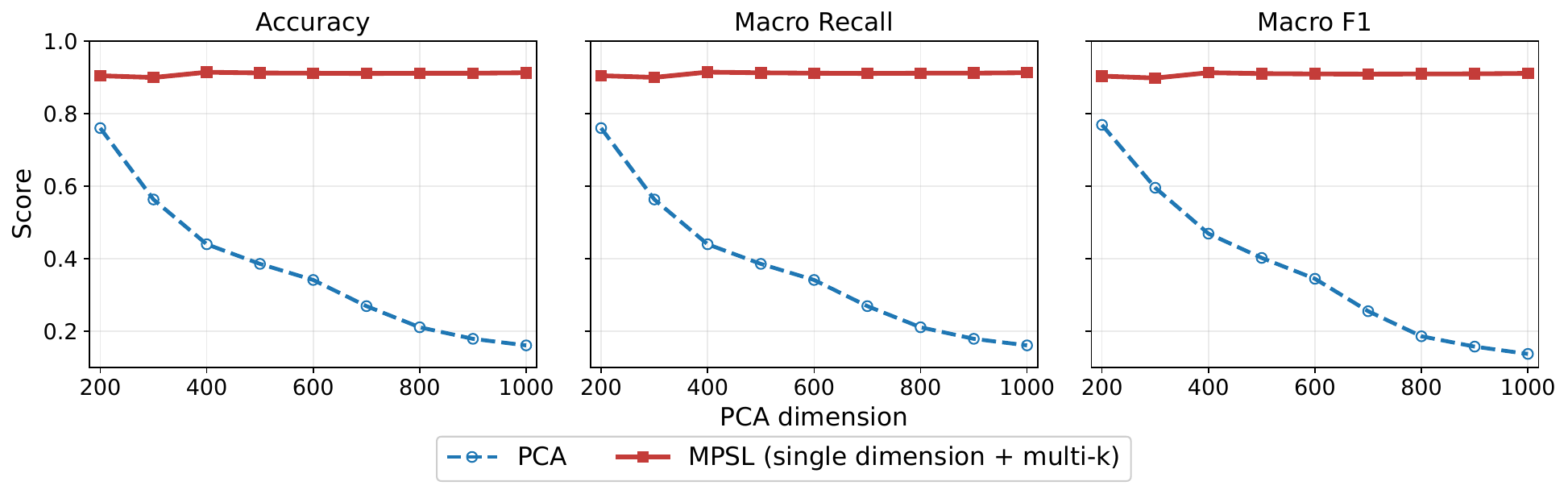}
\caption{COIL20 performance comparison between PCA and MPSL (single dimension $+$ multi-$k$) over PCA dimensions 200--1000, evaluated using Accuracy (Acc), Macro Recall (MR), and Macro F1.}
\label{fig:coil20_pca_vs_psl_one}
\end{figure}

Figure~\ref{fig:coil20_pca_vs_psl_one} illustrates these performance trends across PCA dimensions.
The figure highlights the strong dependence of PCA on dimension selection and the relative stability of MPSL under the same classifier and evaluation protocol.

\begin{figure}[ht]
\centering
\includegraphics[width=\linewidth]{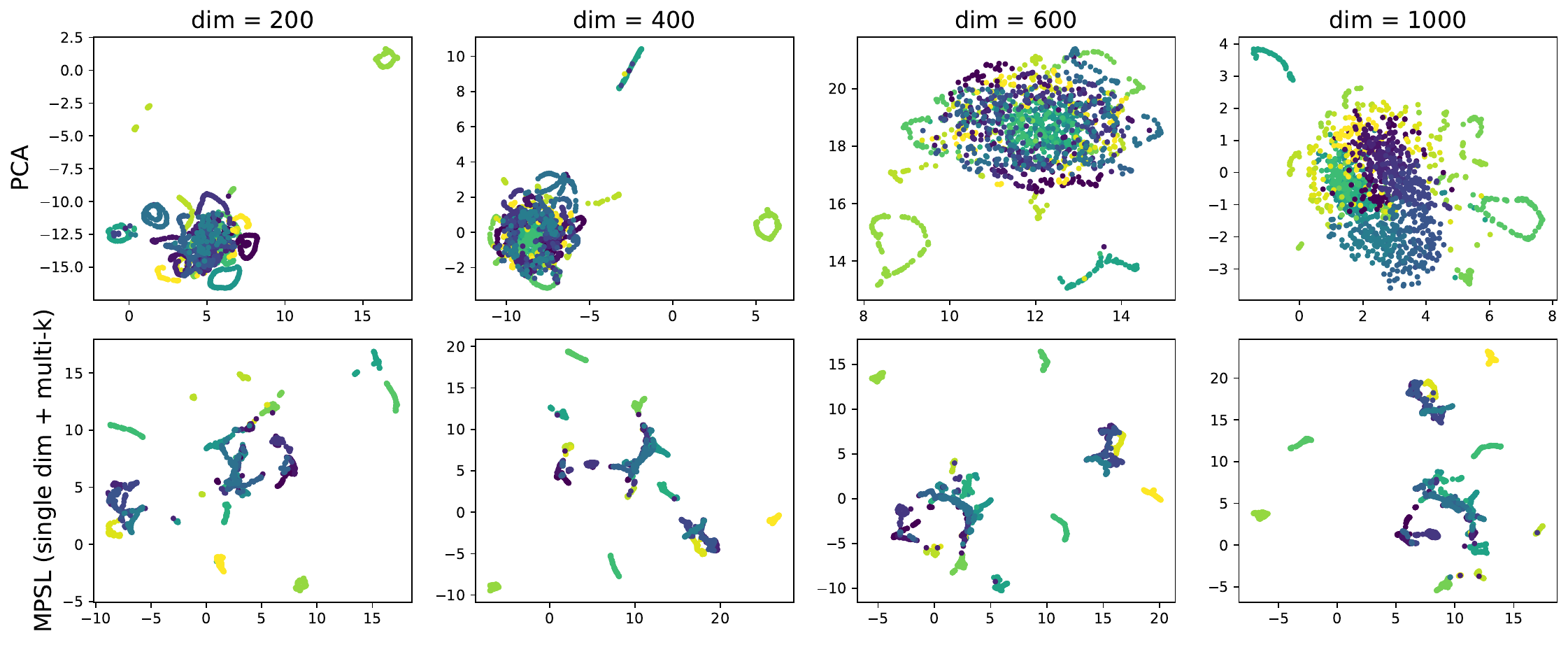}
\caption{
    Qualitative UMAP visualizations of PCA-based and MPSL-based representations on the COIL20 dataset under different reduced dimensions.
    All embeddings are produced using the default UMAP parameters and are intended solely for qualitative visualization.
    The top row shows embeddings of PCA-based features at different reduced dimensions, while the bottom row shows MPSL representations constructed by aggregating single-dimension features across multiple neighborhood scales.
    UMAP is used here to highlight the geometric organization and structural consistency of the learned representations rather than to optimize separability.
}
\label{fig:coil20_umap}
\end{figure}

To further illustrate why MPSL exhibits stable performance across different PCA dimensions, we visualize the learned representations using UMAP.
Figure~\ref{fig:coil20_umap} presents qualitative comparisons between PCA-based and MPSL-based features under the same visualization pipeline.
All UMAP embeddings are generated using the default settings of the UMAP algorithm to ensure consistency across different reduced dimensions and feature constructions.

As shown in Figure~\ref{fig:coil20_umap}, PCA-based representations (top row) yield scattered embeddings with noticeable class overlap, and their geometric structures vary substantially as the reduced dimension changes.
In contrast, MPSL representations (bottom row) exhibit more coherent and continuous geometric structures that remain visually stable across different PCA dimensions.
This qualitative robustness provides geometric insight into the dimension-insensitive classification performance of MPSL observed in Figure~\ref{fig:coil20_pca_vs_psl_one}.

While MPSL computed at a single fixed dimension already demonstrates strong robustness to the choice of PCA dimension, it is natural to ask whether further gains can be obtained by integrating information across multiple reduced dimensions.
To this end, we construct an aggregated MPSL representation by combining MPSL spectral features computed over PCA dimensions 200--1000 and multiple neighborhood sizes.
We refer to this setting as MPSL (multi-dimension, multi-$k$ aggregation).

Table~\ref{tab:coil20_psl_summary} summarizes the average performance of the PCA baseline, MPSL (single dimension $+$ multi-$k$), and MPSL (multi-dimension, multi-$k$ aggregation) over PCA dimensions 200--1000.
Compared with PCA, MPSL computed at a single dimension already provides a substantial performance improvement, and aggregating MPSL representations across multiple dimensions yields a further consistent gain across all evaluation metrics.

\begin{table}[t]
\centering
\small
\setlength{\tabcolsep}{8pt}
\renewcommand{\arraystretch}{1.15}
\caption{Summary performance on COIL20 over PCA dimensions 200--1000. 
Average results are reported for PCA and MPSL (single dimension $+$ multi-$k$), 
while MPSL (multi-dimension, multi-$k$ aggregation) corresponds to a single aggregated representation.}
\label{tab:coil20_psl_summary}
\begin{tabular}{lccc}
\hline
Method & Acc & MR & Macro F1 \\
\hline
PCA (average) 
& 0.3675 & 0.3675 & 0.3682 \\
MPSL (single dimension + multi-$k$, average)
& 0.9091 & 0.9093 & 0.9074 \\
MPSL (multi dimension, multi-$k$ aggregation)
& \textbf{0.9167} & \textbf{0.9169} & \textbf{0.9149} \\
\hline
\end{tabular}
\end{table}

\begin{figure}[ht]
\centering
\includegraphics[width=\linewidth]{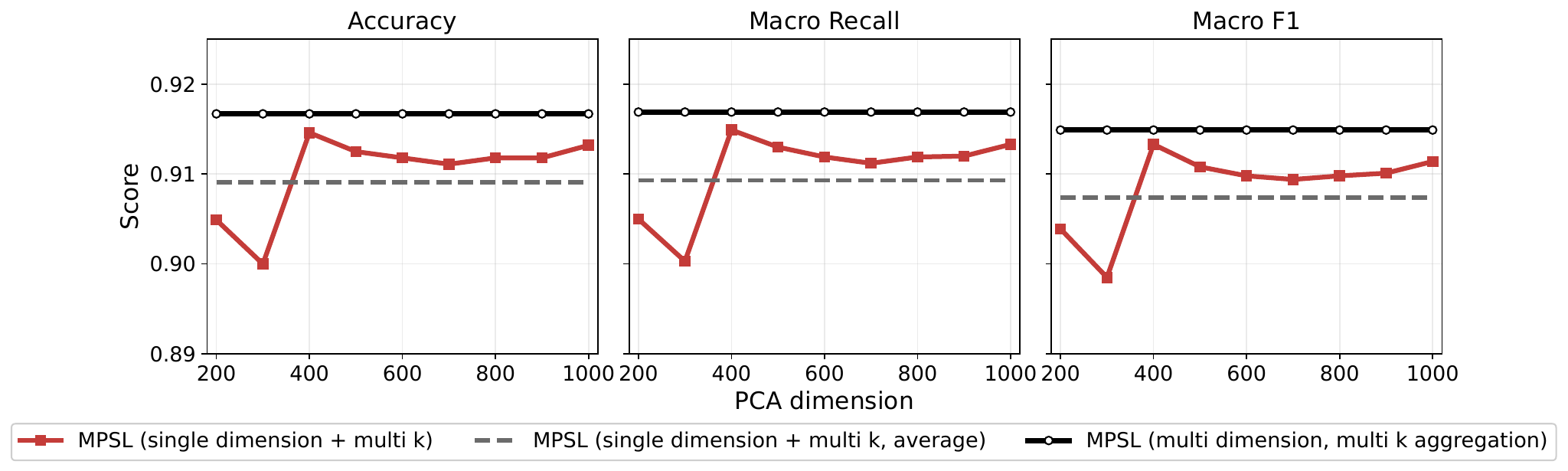}
\caption{Comparison between MPSL (single dimension $+$ multi-$k$) and MPSL (multi-dimension, multi-$k$ aggregation) on COIL20.
Dashed lines denote the average performance of single-dimension MPSL, while solid lines indicate the aggregated multi-dimension MPSL result.}
\label{fig:coil20_psl_one_vs_multi}
\end{figure}

Figure~\ref{fig:coil20_psl_one_vs_multi} further compares MPSL (single dimension $+$ multi-$k$) with the aggregated MPSL representation.
For each metric, the dashed horizontal line indicates the average performance of MPSL computed at a single dimension, while the solid horizontal line corresponds to the multi-dimension aggregated MPSL result.
Across all three metrics, the aggregated representation consistently matches or slightly improves upon the best single-dimension results, indicating that integrating information across dimensions provides a modest but systematic performance gain.

Overall, these results show that MPSL provides a robust alternative to PCA in high-dimensional settings.
While MPSL computed at a single dimension reduces dimension-selection sensitivity, aggregating representations across multiple dimensions leads to a modest but consistent performance improvement.

\subsubsection{ETH80}
\label{subsubsec:eth80_results}

We next evaluate the proposed method on the ETH80 dataset using the same experimental settings as in the COIL20 experiments. Table~\ref{tab:eth80_pca_vs_psl_one} reports classification results on ETH80 in terms of Accuracy (Acc), Macro Recall (MR), and Macro F1.

\begin{table*}[ht]
\centering
\small
\setlength{\tabcolsep}{5pt}
\renewcommand{\arraystretch}{1.15}
\caption{ETH80 results for PCA and MPSL (single dimension $+$ multi-$k$) over PCA dimensions 200 to 1000. Metrics are Accuracy (Acc), Macro Recall (MR), and Macro F1.}
\label{tab:eth80_pca_vs_psl_one}
\begin{tabular}{c|ccc|ccc}
\hline
PCA dim
& \multicolumn{3}{c|}{PCA}
& \multicolumn{3}{c}{MPSL } \\
\cline{2-7}
& Acc & MR & Macro F1
& Acc & MR & Macro F1 \\
\hline
200  & 0.6061 & 0.6061 & 0.5876 & 0.6162 & 0.6162 & 0.6052 \\
300  & 0.4817 & 0.4817 & 0.4370 & 0.6256 & 0.6256 & 0.6122 \\
400  & 0.4149 & 0.4149 & 0.3520 & 0.6305 & 0.6305 & 0.6218 \\
500  & 0.3713 & 0.3713 & 0.3106 & 0.6256 & 0.6256 & 0.6160 \\
600  & 0.3326 & 0.3326 & 0.2866 & 0.6345 & 0.6345 & 0.6242 \\
700  & 0.2963 & 0.2963 & 0.2573 & 0.6357 & 0.6357 & 0.6263 \\
800  & 0.2646 & 0.2646 & 0.2290 & 0.6360 & 0.6360 & 0.6263 \\
900  & 0.2320 & 0.2320 & 0.1939 & 0.6381 & 0.6381 & 0.6282 \\
1000 & 0.2190 & 0.2190 & 0.1781 & 0.6332 & 0.6332 & 0.6237 \\
\hline
Average
& 0.3576 & 0.3576 & 0.3147
& 0.6306 & 0.6306 & 0.6204 \\
\hline
\end{tabular}
\end{table*}

As shown in Table~\ref{tab:eth80_pca_vs_psl_one}, the performance of the PCA baseline decreases steadily as the reduced dimension increases from 200 to 1000.
In contrast, MPSL (single dimension $+$ multi-$k$) consistently achieves higher classification performance across all tested dimensions, with relatively small variation.
These results indicate that the MPSL representation is less sensitive to the choice of reduced dimension on ETH80.

\begin{figure}[ht]
\centering
\includegraphics[width=\linewidth]{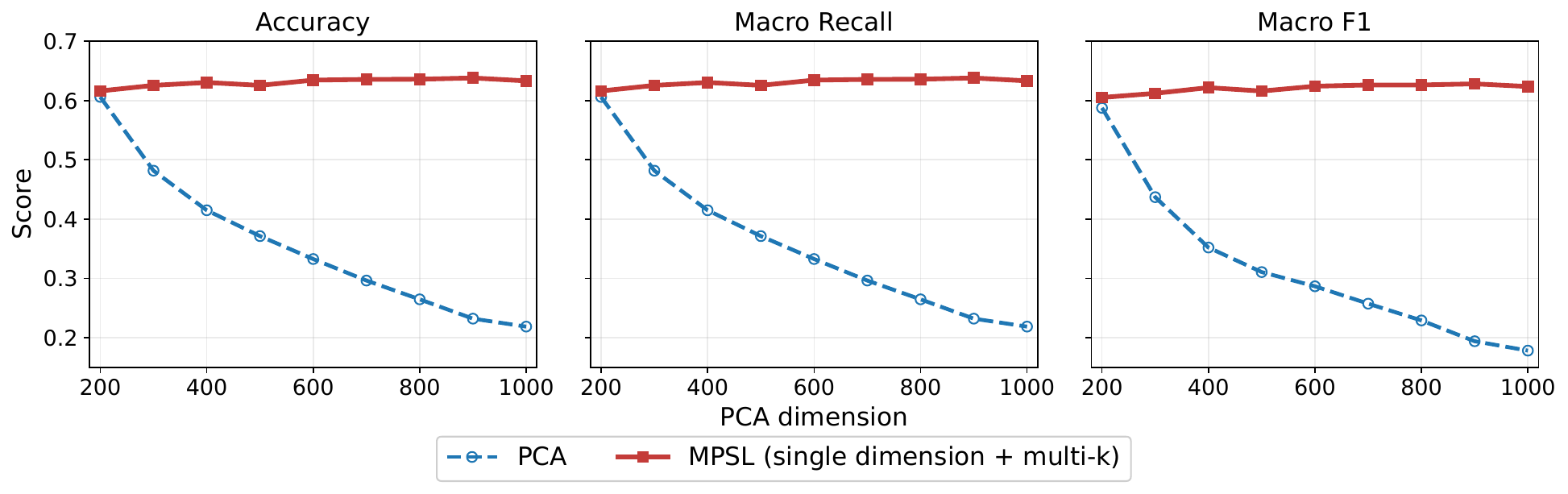}
\caption{ETH80 performance comparison between PCA and MPSL (single dimension $+$ multi-$k$) over PCA dimensions 200 to 1000, evaluated using Accuracy (Acc), Macro Recall (MR), and Macro F1.}
\label{fig:eth80_pca_vs_psl_one}
\end{figure}

Figure~\ref{fig:eth80_pca_vs_psl_one} visualizes these trends across the three evaluation metrics.
While the PCA baseline exhibits a clear dependence on the selected dimension, the MPSL curves remain comparatively stable over the entire range.

We further investigate whether aggregating information across multiple reduced dimensions can provide additional benefits.
To this end, we construct an aggregated MPSL representation by combining spectral features computed over PCA dimensions 200 to 1000 and multiple neighborhood sizes.
This setting is referred to as MPSL (multi-dimension, multi-$k$ aggregation).

\begin{table}[t]
\centering
\small
\setlength{\tabcolsep}{6pt}
\renewcommand{\arraystretch}{1.15}
\caption{Summary performance on ETH80 over PCA dimensions 200--1000. Average results are reported for PCA and MPSL (single dimension $+$ multi-$k$), while MPSL (multi-dimension, multi-$k$ aggregation) corresponds to a single aggregated representation.}
\label{tab:eth80_summary_200_1000}
\begin{tabular}{lccc}
\hline
Method & Acc & MR & Macro F1 \\
\hline
PCA (average) & 0.3576 & 0.3576 & 0.3147 \\
MPSL (single dimension $+$ multi-$k$, average) & 0.6306 & 0.6306 & 0.6204 \\
MPSL (multi-dimension, multi-$k$ aggregation) & \textbf{0.6622} & \textbf{0.6622} & \textbf{0.6536} \\
\hline
\end{tabular}
\end{table}

As summarized in Table~\ref{tab:eth80_summary_200_1000}, aggregating MPSL representations across multiple dimensions leads to a consistent improvement over the average single-dimension MPSL results.
This suggests that integrating spectral information across dimensions captures complementary structure that is not fully reflected at any single reduced dimension.

\begin{figure}[hbpt!]
\centering
\includegraphics[width=\linewidth]{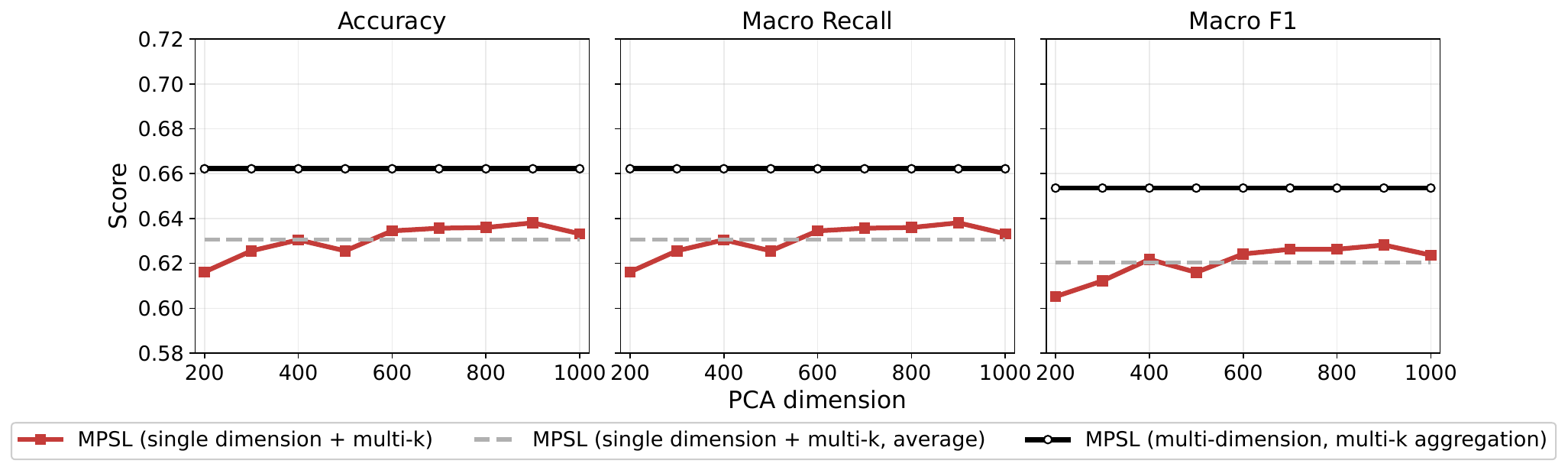}
\caption{ETH80 comparison between MPSL (single dimension $+$ multi-$k$) and MPSL (multi-dimension, multi-$k$ aggregation). Dashed lines indicate the average performance of MPSL (single dimension), while solid reference lines correspond to the aggregated MPSL representation.}
\label{fig:eth80_psl_one_vs_multi}
\end{figure}

Figure~\ref{fig:eth80_psl_one_vs_multi} further illustrates this effect by comparing the single-dimension MPSL results across PCA dimensions with the aggregated MPSL representation.
Across all three metrics, the aggregated MPSL achieves higher performance than the average single-dimension results.

Overall, the results on ETH80 show that MPSL maintains stable performance across a wide range of reduced dimensions on a dataset that is substantially larger than COIL20.
Although the absolute performance gains over PCA are more moderate, MPSL consistently outperforms the PCA baseline at all tested dimensions and exhibits reduced sensitivity to dimension selection.
The multi-dimension aggregation further improves upon the average single-dimension results, indicating that the proposed framework remains effective and stable when applied to larger-scale image datasets.

\subsection{Result summary}

Compared with COIL20, ETH80 contains multiple object instances per class and exhibits larger intra-class variability, making it a more challenging benchmark for image classification.
On COIL20, the proposed MPSL framework achieves consistently high classification performance and shows strong stability across a wide range of reduced dimensions.
In particular, even the single-dimension MPSL representation yields robust results, indicating that MPSL is substantially less sensitive to the choice of reduced dimension than PCA.

On ETH80, although the absolute classification accuracy is lower than that observed on COIL20, MPSL continues to demonstrate stable performance across reduced dimensions and consistently outperforms the PCA baseline.
It is worth noting that ETH80 is more than twice the size of COIL20, while the neighborhood sizes used in the MPSL construction are kept identical across the two datasets.
This suggests that the local neighborhood parameters may not be optimal for the larger and more heterogeneous ETH80 dataset.
Motivated by this observation, we further investigate the effect of increasing neighborhood sizes in the MPSL construction, and discuss the corresponding results in the Discussion section.

Overall, these experiments show that MPSL is effective in high-dimensional regimes where PCA representations become increasingly noisy.
The results from the single-dimension setting already indicate that MPSL exhibits low sensitivity to the choice of reduced dimension, while the multi-dimension aggregation further improves performance by integrating complementary information across scales.

\section{Discussion}
\label{sec:discussion}

The experimental results demonstrate that the proposed multi-dimensional persistent sheaf Laplacian (MPSL) framework provides stable and consistent performance across a wide range of reduced dimensions on both COIL20 and ETH80. While COIL20 achieves particularly high classification accuracy overall, ETH80 represents a more challenging setting due to its larger scale and increased intra-class variability arising from multiple object instances per category.

It is worth noting that PCA-based representations can attain relatively high classification accuracy at very small reduced dimensions (e.g., below 100), particularly on COIL20. However, such performance is highly sensitive to the specific choice of dimension and degrades rapidly as the dimensionality increases. In this work, our primary objective is not to identify an optimal reduced dimension for PCA, but to evaluate the robustness of representation learning methods under varying dimensional settings. From this perspective, the observed peak performance of PCA at small dimensions highlights its strong dependence on careful dimension tuning, rather than contradicting the motivation of the proposed framework.

Furthermore, when informative PCA dimensions (such as $d=20$) are incorporated within the multi-dimensional, multi-$k$ aggregation framework, the proposed MPSL representation achieves further performance improvement, while the single-dimension variant remains consistently stable across the entire dimensional range. This indicates that the framework does not discard favorable dimensional settings, but effectively integrates them into a more robust representation. For completeness, we report classification results of COIL20 over the full range of PCA dimensions from 10 to 1000 in the Supporting Information, Section~3.

In the main experiments, we intentionally adopt the same range of neighborhood sizes for ETH80 as for COIL20 in order to maintain a consistent and fair experimental protocol. Under this setting, MPSL already yields a substantial improvement over PCA and exhibits markedly reduced sensitivity to the choice of reduced dimension. However, the summary results also suggest that ETH80 retains additional room for improvement through the incorporation of broader local context.

To further investigate this effect, we examine the impact of increasing the maximum neighborhood size in the multi-$k$ aggregation, with results reported in the Supporting Information, Section 4. Across Accuracy, Macro Recall, and Macro F1, the performance curves show consistent and gradual improvements as the neighborhood size increases, without introducing noticeable instability. This trend aligns with the structural characteristics of ETH80, where larger neighborhoods can
capture more intra-class variation among object instances.

These observations highlight an important property of the proposed framework. MPSL does not rely on a narrowly tuned neighborhood parameter, but instead benefits from integrating information across multiple neighborhood scales. Even when applied with conservative parameter choices, the method remains robust to high-dimensional noise introduced by large PCA dimensions, while maintaining flexibility to adapt to more complex datasets through controlled expansion of local neighborhoods.

\section{Conclusion}
\label{sec:conclusion}
In this work, we proposed a multi-dimensional  persistent sheaf Laplacian (MPSL) framework for image analysis, with a particular focus on addressing the sensitivity of dimension reduction methods to the choice of reduced dimension. Rather than seeking an optimal embedding at a single dimension, our approach treats different reduced dimensions and neighborhood sizes as complementary scales and integrates their associated spectral information into a unified representation.

A key aspect of the proposed framework is the use of dataset-level simplicial complexes, on which persistent sheaf Laplacians are constructed. Unlike pixel-based or grid-based representations, this construction does not rely on the spatial structure of individual images. As a result, the framework is more general and can be applied to a wide range of data types that admit vector representations and neighborhood relations, beyond standard image grids.

Through experiments on COIL20 and ETH80, we demonstrated that the proposed MPSL representations exhibit strong stability with respect to the choice of reduced dimension, particularly in high-dimensional regimes where PCA-based baselines are highly sensitive. The results show that MPSL computed at a single dimension already provides robust performance, and that integrating information across multiple dimensions can further improve classification accuracy in a consistent manner. Importantly, these gains are achieved without tuning a single optimal dimension or neighborhood parameter.

The goal of this work is not to claim a universally optimal image classification method, but to introduce a robust and stable spectral framework for analyzing high-dimensional data under dimension reduction. By combining simplicial constructions, sheaf Laplacians, and multi-dimensional aggregation, the proposed approach offers a principled alternative to single-scale representations and provides a foundation for further extensions to other data modalities, such as spatial transcriptomic data analysis and other learning tasks, such as clustering and regression.

\section*{Data and Code Availability}

All source code used in this work is publicly available at \url{https://github.com/XiangXiangJY/MPSL}.
The experiments were conducted on two publicly available benchmark datasets:
COIL20, available at \url{https://www.cs.columbia.edu/CAVE/databases/SLAM_coil-20_coil-100/coil-20/coil-20-proc.zip}, 
and ETH80, downloaded from \url{https://github.com/chenchkx/ETH-80}.

 \section*{Supporting Information}

Additional materials  and detailed definitions of the evaluation metrics  are provided in the
Supporting Information.

\section*{Acknowledgments}
This work was supported in part by NIH grant R35GM148196, National Science Foundation grant DMS2052983,  Michigan State University Research Foundation, and  Bristol-Myers Squibb 65109.

\bibliographystyle{plain}
\bibliography{refs}

@article{cang2017topologynet,
  title={TopologyNet: Topology based deep convolutional and multi-task neural networks for biomolecular property predictions},
  author={Cang, Zixuan and Wei, Guo-Wei},
  journal={PLoS computational biology},
  volume={13},
  number={7},
  pages={e1005690},
  year={2017},
  publisher={Public Library of Science San Francisco, CA USA}
}

@article{papamarkou2024position,
  title={Position: Topological deep learning is the new frontier for relational learning},
  author={Papamarkou, Theodore and Birdal, Tolga and Bronstein, Michael and Carlsson, Gunnar and Curry, Justin and Gao, Yue and Hajij, Mustafa and Kwitt, Roland and Lio, Pietro and Di Lorenzo, Paolo and others},
  journal={Proceedings of machine learning research},
  volume={235},
  pages={39529},
  year={2024}
}

@article{lesnick2015interactive,
  title={Interactive visualization of 2-d persistence modules},
  author={Lesnick, Michael and Wright, Matthew},
  journal={arXiv preprint arXiv:1512.00180},
  year={2015}
}

@techreport{nene1996columbia,
  title={Columbia object image library (coil-20)},
  author={Nene, Sameer A and Nayar, Shree K and Murase, Hiroshi and others},
  year={1996},
  institution={Technical report CUCS-005-96}
}

@inproceedings{davies2023persistent,
  title={The persistent Laplacian for data science: Evaluating higher-order persistent spectral representations of data},
  author={Davies, Thomas and Wan, Zhengchao and Sanchez-Garcia, Ruben J},
  booktitle={International Conference on Machine Learning},
  pages={7249--7263},
  year={2023},
  organization={PMLR}
}

@inproceedings{leibe2003analyzing,
  title={Analyzing appearance and contour based methods for object categorization},
  author={Leibe, Bastian and Schiele, Bernt},
  booktitle={2003 IEEE Computer Society Conference on Computer Vision and Pattern Recognition, 2003. Proceedings.},
  volume={2},
  pages={II--409},
  year={2003},
  organization={IEEE}
}

@article{zhou2022pca,
  title={PCA outperforms popular hidden variable inference methods for molecular QTL mapping},
  author={Zhou, Heather J and Li, Lei and Li, Yumei and Li, Wei and Li, Jingyi Jessica},
  journal={Genome biology},
  volume={23},
  number={1},
  pages={210},
  year={2022},
  publisher={Springer}
}

@article{su2025topological,
  title={Topological data analysis and topological deep learning beyond persistent homology: a review},
  author={Su, Zhe and Liu, Xiang and Hamdan, Layal Bou and Maroulas, Vasileios and Wu, Jie and Carlsson, Gunnar and Wei, Guo-Wei},
  journal={Artificial Intelligence Review},
  year={2025},
  publisher={Springer}
}

@article{li2020efficient,
  title={Efficient dimension reduction and surrogate-based sensitivity analysis for expensive models with high-dimensional outputs},
  author={Li, Min and Wang, Ruo-Qian and Jia, Gaofeng},
  journal={Reliability Engineering \& System Safety},
  volume={195},
  pages={106725},
  year={2020},
  publisher={Elsevier}
}

@article{hayes2025persistent,
  title={Persistent Sheaf Laplacian Analysis of Protein Flexibility},
  author={Hayes, Nicole and Wei, Xiaoqi and Feng, Hongsong and Merkurjev, Ekaterina and Wei, Guo-Wei},
  journal={The Journal of Physical Chemistry B},
  volume={129},
  number={17},
  pages={4169--4178},
  year={2025},
  publisher={ACS Publications}
}

@article{liu2025manifold,
  title={Manifold Topological Deep Learning for Biomedical Data},
  author={Liu, Xiang and Su, Zhe and Shi, Yongyi and Tong, Yiying and Wang, Ge and Wei, Guo-Wei},
  journal={arXiv preprint arXiv:2503.00175},
  year={2025}
}

@article{wang2023persistent,
  title={Persistent path laplacian},
  author={Wang, Rui and Wei, Guo-Wei},
  journal={Foundations of data science (Springfield, Mo.)},
  volume={5},
  number={1},
  pages={26},
  year={2023}
}

@article{wei2025persistent,
  title={Persistent topological laplacians—a survey},
  author={Wei, Xiaoqi and Wei, Guo-Wei},
  journal={Mathematics},
  volume={13},
  number={2},
  pages={208},
  year={2025},
  publisher={MDPI}
}

@article{liu2024algebraic,
  title={The Algebraic Stability for Persistent  Laplacians.},
  author={Liu, Jian and Li, Jingyan  and Wu, Jie},
  journal={Homology, Homotopy \& Applications},
  volume={26},
  number={2},
  year={2024}
}

@article{jones2025petls,
  title={PETLS: Persistent Topological Laplacian Software},
  author={Jones, Benjamin and Wei, Guo-Wei},
  journal={arXiv preprint arXiv:2508.11560},
  year={2025}
}

@article{wang2021hermes,
  title={HERMES: Persistent spectral graph software},
  author={Wang, Rui and Zhao, Rundong and Ribando-Gros, Emily and Chen, Jiahui and Tong, Yiying and Wei, Guo-Wei},
  journal={Foundations of data science (Springfield, Mo.)},
  volume={3},
  number={1},
  pages={67},
  year={2021}
}

@article{memoli2022persistent,
  title={Persistent Laplacians: Properties, algorithms and implications},
  author={M{\'e}moli, Facundo and Wan, Zhengchao and Wang, Yusu},
  journal={SIAM Journal on Mathematics of Data Science},
  volume={4},
  number={2},
  pages={858--884},
  year={2022},
  publisher={SIAM}
}

@article{wee2025review,
  title={A review of topological data analysis and topological deep learning in molecular sciences},
  author={Wee, JunJie and Jiang, Jian},
  journal={Journal of Chemical Information and Modeling},
  volume={65},
  number={23},
  pages={12691--12706},
  year={2025},
  publisher={ACS Publications}
}

@inproceedings{zomorodian2004computing,
  title={Computing persistent homology},
  author={Zomorodian, Afra and Carlsson, Gunnar},
  booktitle={Proceedings of the twentieth annual symposium on Computational geometry},
  pages={347--356},
  year={2004}
}

@article{hozumi2022ccp,
  title={{CCP:} Correlated clustering and projection for dimensionality reduction},
  author={Hozumi, Yuta and Wang, Rui and Wei, Guo-Wei},
  journal={arXiv preprint arXiv:2206.04189},
  year={2022}
}

@inproceedings{garin2019topological,
  title={A topological" reading" lesson: Classification of { MNIST using TDA}},
  author={Garin, Ad{\'e}lie and Tauzin, Guillaume},
  booktitle={2019 18th IEEE international conference on machine learning and applications (ICMLA)},
  pages={1551--1556},
  year={2019},
  organization={IEEE}
}

@book{edelsbrunner2010computational,
  title={Computational topology: an introduction},
  author={Edelsbrunner, Herbert and Harer, John},
  year={2010},
  publisher={American Mathematical Soc.}
}

@article{wang2020persistent,
  title={Persistent spectral graph},
  author={Wang, Rui and Nguyen, Duc Duy and Wei, Guo-Wei},
  journal={International journal for numerical methods in biomedical engineering},
  volume={36},
  number={9},
  pages={e3376},
  year={2020},
  publisher={Wiley Online Library}
}

@article{qiu2023persistent,
  title={Persistent spectral theory-guided protein engineering},
  author={Qiu, Yuchi and Wei, Guo-Wei},
  journal={Nature computational science},
  volume={3},
  number={2},
  pages={149--163},
  year={2023},
  publisher={Nature Publishing Group US New York}
}

@article{mallat2012group,
  title={Group invariant scattering},
  author={Mallat, St{\'e}phane},
  journal={Communications on Pure and Applied Mathematics},
  volume={65},
  number={10},
  pages={1331--1398},
  year={2012},
  publisher={Wiley Online Library}
}

@incollection{burt1987laplacian,
  title={The Laplacian pyramid as a compact image code},
  author={Burt, Peter J and Adelson, Edward H},
  booktitle={Readings in computer vision},
  pages={671--679},
  year={1987},
  publisher={Elsevier}
}

@article{burges2010dimension,
  title={Dimension reduction: A guided tour},
  author={Burges, Christopher JC and others},
  journal={Foundations and Trends{\textregistered} in Machine Learning},
  volume={2},
  number={4},
  pages={275--365},
  year={2010},
  publisher={Now Publishers, Inc.}
}

@book{bishop2006pattern,
  title={Pattern recognition and machine learning},
  author={Bishop, Christopher M and Nasrabadi, Nasser M},
  volume={4},
  number={4},
  year={2006},
  publisher={Springer}
}

@book{gonzalez2009digital,
  title={Digital image processing},
  author={Gonzalez, Rafael C},
  year={2009},
  publisher={Pearson education india}
}

@article{fodor2002survey,
  title={A survey of dimension reduction techniques},
  author={Fodor, Imola K and others},
  year={2002},
  publisher={Technical Report UCRL-ID-148494, Lawrence Livermore National Laboratory}
}

@article{mcinnes2018umap,
  title={Umap: Uniform manifold approximation and projection for dimension reduction},
  author={McInnes, Leland and Healy, John and Melville, James},
  journal={arXiv preprint arXiv:1802.03426},
  year={2018}
}

@article{hand2001simple,
  title={A simple generalisation of the area under the ROC curve for multiple class classification problems},
  author={Hand, David J and Till, Robert J},
  journal={Machine learning},
  volume={45},
  number={2},
  pages={171--186},
  year={2001},
  publisher={Springer}
}

@article{sokolova2009systematic,
  title={A systematic analysis of performance measures for classification tasks},
  author={Sokolova, Marina and Lapalme, Guy},
  journal={Information processing \& management},
  volume={45},
  number={4},
  pages={427--437},
  year={2009},
  publisher={Elsevier}
}

@inproceedings{lewis1991evaluating,
  title={Evaluating text categorization i},
  author={Lewis, David D},
  booktitle={Speech and Natural Language: Proceedings of a Workshop Held at Pacific Grove, California, February 19-22, 1991},
  year={1991}
}

@article{hansen2019toward,
  title={Toward a spectral theory of cellular sheaves},
  author={Hansen, Jakob and Ghrist, Robert},
  journal={Journal of Applied and Computational Topology},
  volume={3},
  number={4},
  pages={315--358},
  year={2019},
  publisher={Springer}
}

@book{curry2014sheaves,
  title={Sheaves, cosheaves and applications},
  author={Curry, Justin Michael},
  year={2014},
  publisher={University of Pennsylvania}
}

@book{lundell2012topology,
  title={The topology of CW complexes},
  author={Lundell, Albert T and Weingram, Stephen},
  year={2012},
  publisher={Springer Science \& Business Media}
}

@article{bro2014principal,
  title={Principal component analysis},
  author={Bro, Rasmus and Smilde, Age K},
  journal={Analytical methods},
  volume={6},
  number={9},
  pages={2812--2831},
  year={2014},
  publisher={Royal society of chemistry}
}

@article{maaten2008visualizing,
  title={Visualizing data using t-SNE},
  author={Maaten, Laurens van der and Hinton, Geoffrey},
  journal={Journal of machine learning research},
  volume={9},
  number={Nov},
  pages={2579--2605},
  year={2008}
}

@article{lee1999learning,
  title={Learning the parts of objects by non-negative matrix factorization},
  author={Lee, Daniel D and Seung, H Sebastian},
  journal={nature},
  volume={401},
  number={6755},
  pages={788--791},
  year={1999},
  publisher={Nature Publishing Group UK London}
}
   
\end{document}